\documentclass[10pt,
final,
 journal, 
twocolumn,
 compsoc,
 ]{IEEEtran} 
\usepackage{spconf,amsmath,graphicx}

\usepackage{amsmath}
\usepackage{amssymb}
\usepackage{amsfonts}
\usepackage{xcolor}
\usepackage{graphics}
\usepackage{subcaption}
\usepackage{dblfloatfix}
\usepackage[utf8]{inputenc}
\usepackage{graphics}
\usepackage{epsfig}
\usepackage{caption}
\usepackage{algorithm}
\usepackage{algorithmic}

\title{Two-way Spectrum Pursuit for $\boldsymbol{CUR}$ Decomposition and Its Application in Joint Column/Row Subset Selection}

\begin{document}
\maketitle
\vspace{-2mm}
\begin{abstract}
The problem of simultaneous column and row subset selection is addressed in this paper. The column space and row space of a matrix are spanned by its left and right singular vectors, respectively. However, the singular vectors are not within actual columns/rows of the matrix. In this paper, an iterative approach is proposed to capture the  most structural information of  columns/rows  via selecting a subset of actual columns/rows. This algorithm is referred to as two-way spectrum pursuit (TWSP) which provides us with an accurate solution for the CUR matrix decomposition. TWSP is applicable in a wide range of applications since it enjoys a linear complexity w.r.t. number of original columns/rows. We demonstrated the application of TWSP for joint channel and sensor selection in cognitive radio networks, informative users and contents detection, and efficient supervised data reduction. 
\end{abstract}
\vspace{-2mm}
\section{Introduction}
\vspace{-1mm}
 
\begin{figure*}[h!]
    \centering
    \includegraphics[width=0.83\textwidth]{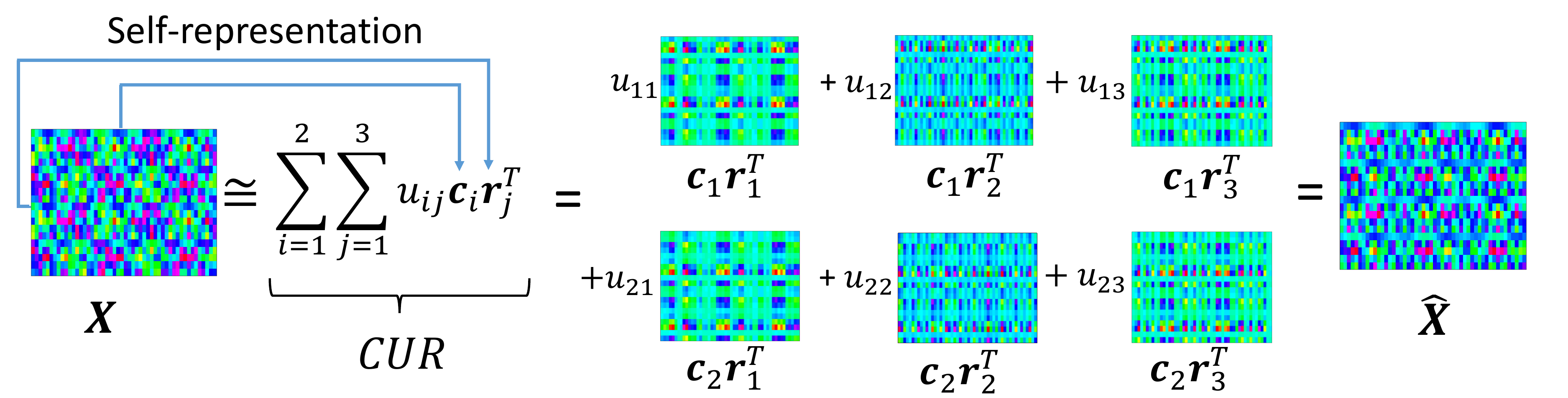}
    \vspace{-3mm}
    \caption{\small{Two columns and three rows from matrix $\boldsymbol{X}$ are selected and organized in matrix $\boldsymbol{C}$ and $\boldsymbol{R}$. The outer product of each pair of  a selected column and a selected row constructs a rank-1 matrix, i.e., $\boldsymbol{c}_i\boldsymbol{r}_j^T$. The  contribution amount of each pair is reflected in variable $u_{ij}$. The core matrix $\boldsymbol{U}$ is the collection of all $u_{ij}$'s. The goal is to minimize $\|\boldsymbol{X}-\hat{\boldsymbol{X}}\|_F$ where $\hat{\boldsymbol{X}}=\boldsymbol{CUR}$. } }
    \vspace{-4mm}
    \label{fig:unfold}
\end{figure*}

 Matrix factorization provides a concise representation of data. Despite desirable uniqueness conditions and computational simplicity of the well-known singular value decomposition (SVD), it comes with some fundamental shortcomings. The intrinsic structure of data is not inherited to the singular components.  Moreover, SVD implies orthogonality  on the components which is irrelevant to the underlying structure of the original data. This enforced  structure makes the bases, a.k.a. singular vectors, hard to interpret \cite{hamm2020perspectives}. On the other hand, it is shown that borrowing bases from the actual samples of a dataset provides a robust representation, which can be employed in interesting applications where sampling is  their key factor \cite{wang2013improving}. This problem is studied under the literature of column subset selection problem (CSSP) \cite{deshpande2010efficient,boutsidis2009improved} and CUR decomposition \cite{mahoney2009cur,boutsidis2017optimal}. A general problem for CSSP and CUR can be written in the following form:
\begin{equation}
\label{eq:main}
\small
    (\mathbb{S}^c,\mathbb{S}^r)=\underset{\mathbb{S}^c,\mathbb{S}^r}{\text{argmin}}\|{\boldsymbol{X}}-\pi_{\mathbb{S}^r}^{r}(\pi_{\mathbb{S}^c}^{c}({\boldsymbol{X}}))\|_F^2,
\end{equation}
where, $\!\small{\boldsymbol{X}}\!\!\in\!\! \mathbb{R}^{N\times M}\!$ is the data matrix containing $M$ data points in an $N$-dimensional space. Here, $\pi_{\mathbb{S}^c}^{c}(.)$ and $\pi_{\mathbb{S}^r}^{r}(.)$ indicate column space projection and row space projection, respectively. These operators project all columns (rows) to a low-dimensional subspace spanned by selected columns (rows) of matrix $\boldsymbol{X}$ indexed by the set $\mathbb{S}^c$ ($\mathbb{S}^r$). The chronological order in applying $\pi_{\mathbb{S}^c}^{c}(.)$ and $\pi_{\mathbb{S}^r}^{r}(.)$ does not affect the problem since these operators are linear. Moreover, substituting $\pi_{\mathbb{S}^r}^{r}$ with the identity projection simplifies the problem to CSSP. Fig. \ref{fig:unfold} illustrates the structure of CUR matrix decomposition as a self-representative approach.

A versatile metric for evaluating the performance of a data subset selection algorithm can be defined by the approximation error resulted from the projection of the entire data to the span of selected rows/columns.
How close to the optimal selection an algorithm can reach, is determined by comparing its approximation error to the best low-rank approximation error specified by the spectral decomposition.  Recently, we proposed a fast and accurate algorithm for solving CSSP which is called spectrum pursuit (SP) \cite{joneidi2020select}.
Inspired by SP, we propose a new algorithm to address the more general case of the CUR matrix decomposition. 
The main contributions of our paper are summarized as follows:
\vspace{-2mm}
\begin{itemize}
\small
    \item A novel algorithm for CUR  decomposition, referred to as two-way spectrum pursuit (TWSP), is proposed. TWSP provides an accurate solution for CUR decomposition. 
    \item TWSP  enjoys a linear complexity w.r.t. the number of columns and the number of rows of a matrix.
    \item TWSP is a parameter-free algorithm that only requires the number of desired columns and rows for selection. Thus, TWSP does not require any parameter fine-tuning.
    \item The  TWSP algorithm is put to the test and investigated in a set of synthetic and real experiments.
    \item The role of the core matrix $\boldsymbol{U}$ in $\boldsymbol{CUR}$ decomposition is illustrated which shows the connection between selected columns and rows. Based on analysis of $\boldsymbol{U}$, an interesting application for joint sensor/channel selection is presented. 
\end{itemize}

\vspace{-2mm}
\textbf{Notations:} Throughout this paper, vectors, and matrices are denoted by bold lowercase, and bold uppercase letters, respectively. Moreover sets are indicated by blackboard bold characters. The $m^{\text{th}}$ column of matrix $\boldsymbol{X}$ is denoted by $\boldsymbol{X}(:,m)$ and the $n^{\text{th}}$ row of matrix $\boldsymbol{X}$ is denoted by $\boldsymbol{X}(n,:)$. 

The rest of paper is organized as follows. Sec. \ref{sec:background} studies the problem of column subset selection as the conventional one-way selection scheme. In Sec. \ref{sec:twsp}, our main work is presented for joint column and row subset selection. Experimental results are exhibited in Sec. \ref{sec:exp}.
\section{Column Subset Selection Problem}


\label{sec:background}
Selecting the most diverse subset of data in an optimal sense is studied vastly \cite{deshpande2010efficient,li2017polynomial,joneidi2018optimal}. However, these methods do not guarantee that the unselected columns are  \emph{well represented} by the selected ones. Further, outliers are selected with a high probability using such algorithms due to their diversity \cite{joneidi2020select}. 
 A more effective approach is selecting some \emph{representatives} which are able to approximate the rest of data accurately \cite{elhamifar2012see} as defined as a special case of \eqref{eq:main}. 
This is an NP-hard problem \cite{shitov2020column} and there are several efforts for solving this  problem \cite{boutsidis2014near,deshpande2010efficient,paul2015column,zaeemzadeh2019iterative}. There are computationally expensive approaches based on convex relaxation \cite{elhamifar2012see,elhamifar2015dissimilarity} that are not computationally feasible for large datasets since their complexity is of order $O(M^3)$, where $M$ is the number of original columns. Recently, we proposed a new algorithm for solving CSSP with a linear complexity which is called Spectrum pursuit (SP)~\cite{joneidi2020select}. The SP algorithm finds $K$ columns of $\boldsymbol{X}$ such that their span is close to that of the best rank-$K$ approximation of $\boldsymbol{X}$. SP is an iterative approach where at each iteration one sample selection is optimized such that the ensemble of selected samples describes the whole dataset more accurately.  SP finds representatives such that the column space is spanned accurately via consecutive rank-$1$ approximations as theoretically analyzed in \cite{joneidi2020optimality}. In the present paper, we extend the SP algorithm for selecting columns and rows jointly such that their outer product can represent the whole matrix accurately. A naive approach is applying SP algorithm on the matrix of interest to select a subset of columns and applying SP on its transpose in order to find a subset of rows. However, this approach is not efficient and we will compare it with our proposed approach which is optimized  through a joint representation of selected columns and rows.

\vspace{-3mm}
\section{Two-way Spectrum Pursuit }
\label{sec:twsp}
 The introduced joint column/row subset selection in \eqref{eq:main} can be written as a CUR decomposition in the following form in which factor matrices must be drawn from actual columns/rows of the original matrix as
\begin{align}
\small
\label{eq:problem_equ}
  (\boldsymbol{C},\boldsymbol{U},\boldsymbol{R})=\underset{\boldsymbol{C},\boldsymbol{U},\boldsymbol{R}}{\text{argmin}}\|{\boldsymbol{X}}-\boldsymbol{C}\boldsymbol{U}\boldsymbol{R}\|_F^2 , \\
  \text{s.t.}\;\;\boldsymbol{c}_k \in \mathbb{X}_c \;\text{and}\;\boldsymbol{r}_k\in\mathbb{X}_r.\nonumber
\end{align}
In this problem, $\mathbb{X}_c$, and $\mathbb{X}_r$ indicate the set of normalized columns and rows of matrix $\boldsymbol{X}$, respectively. Here, $\boldsymbol{c}_k$ and $\boldsymbol{r}^T_k$ denote the $k^{th}$ column and the $k^{th}$ row of $\boldsymbol{C}$ and $\boldsymbol{R}$, respectively. In other words, $\mathbb{X}_c\!=\!\{\boldsymbol{X}(:,m)/\|\boldsymbol{X}(:,m)\|\}$ for all columns and  $\mathbb{X}_r\!=\!\{\boldsymbol{X}(n,:)/\|\boldsymbol{X}(n,:)\|\}$ for all rows. 
Please note that replacing constraints in \eqref{eq:problem_equ} with orthogonality constraint on $\boldsymbol{c}_k$'s and on $\boldsymbol{r}_k$'s results in the truncated singular value decomposition (SVD) with $K$ most significant components. In this case, $\boldsymbol{C}$  and $\boldsymbol{R}$ contain the first $K$ left singular vectors and the first $K$ right singular vectors of $\boldsymbol{X}$, respectively. Moreover, the core matrix will be diagonal and the entries will be singular values with diagonal entries as singular values. However, the underlying constraints in (\ref{eq:problem_equ}) turn the problem into a joint subset of row and column selection problem instead of matrix low-rank approximation problem.  

To solve this complicated problem, we split it into two consecutive problems for optimization on the $k^{\text{th}}$ selected column/row. Our optimization approach is alternative, i.e., a random subset of columns and rows are picked. Then, one column or row is considered to be replaced with a more efficient one at each iteration. Since scaling  a vector does not change its span, without loss of generality we assume that the column or the row subject of the optimization lie on the unit sphere. At each iteration, a rank-$1$ component is optimized characterized by $\boldsymbol{c}\boldsymbol{g}^{T}$ or $\boldsymbol{h}\boldsymbol{r}^{T}$ given by 

\begin{subequations}
\small
\vspace{-1mm}
\label{eq:seperated}
\begin{equation}
\label{eq:problem_sep1}
  \underset{\boldsymbol{c},\;\boldsymbol{W},\;\boldsymbol{g}}{\text{argmin}}\;\;\|\underbrace{\boldsymbol{X}-\boldsymbol{C}_k\boldsymbol{W}\boldsymbol{R}}_{\boldsymbol{E}^c}- \boldsymbol{c}\boldsymbol{g}^T\|_F^2 \;\text{s.t.}\;\;\|\boldsymbol{c}\|_2=1 
  \end{equation}
  \begin{equation}
  \label{eq:problem_sep2}
   \underset{\boldsymbol{h},\;\boldsymbol{Y},\;\boldsymbol{r}}{\text{argmin}}\;\;\|\underbrace{\boldsymbol{X}-\boldsymbol{C}\boldsymbol{Y}\boldsymbol{R}_k}_{\boldsymbol{E}^r}- \boldsymbol{h}\boldsymbol{r}^T\|_F^2 \;\text{s.t.}\;\;\|\boldsymbol{r}\|_2=1
    \end{equation}
\end{subequations}

Matrix $\boldsymbol{C}_k$ is the set of selected columns except the $k^{\text{th}}$ one and $\boldsymbol{R}_k$ is the set of selected rows except the $k^{\text{th}}$ row.  The first subproblem can be solved easily w.r.t. $\boldsymbol{c}$ using singular value decomposition. In other words, $\boldsymbol{cg}^T$ and $\boldsymbol{hr}^T$ are the best rank-$1$ approximations of the residual $\boldsymbol{E}^c$ and $\boldsymbol{E}^r$, respectively. The obtained $\boldsymbol{c}$/$\boldsymbol{r}$ is the best column/row that can be added to the pool of selected columns/rows. However, the obtained vector is not available in the given dataset as a column or row since it is a singular vector which is a function of all columns/rows.  The following step re-imposes the underlying constraints at each iteration,

\begin{subequations}
\vspace{-4mm}
\small
\label{eq:matched}
\begin{equation}
\label{eq:problem_match1}
\mathbb{S}_k^c=\underset{m}{\text{argmax}}|\boldsymbol{x}_m^T\boldsymbol{c}|,\;\; \forall{\boldsymbol{x}_m}\in \mathbb{X}_c
  \end{equation}
  \begin{equation}
  \label{eq:problem_match2}
\mathbb{S}_k^r=\underset{n}{\text{argmax}}|\boldsymbol{x}_n^T\boldsymbol{r}|,\;\;\forall{\boldsymbol{x}_n}\in \mathbb{X}_r
    \end{equation}
\end{subequations}
Here, $\mathbb{S}_k^c$ indicates a singleton that contains $k^{\text{th}}$ selected column and $\mathbb{S}_k^r$ corresponds to the $k^{\text{th}}$ selected row. In each iteration, one column or one row is the subject of optimization. The impact of the latest estimation for that column/row on the representation is neglected. Then, an optimized replacement is found. 
At each iteration of TWSP, sub-problems in \eqref{eq:seperated} are solved and their solutions are matched to the accessible column samples and row samples through matching equations in \eqref{eq:matched}. The new selected column or row is stored in $\mathbb{S}_k^c$ or $\mathbb{S}_k^r$, respectively. At each iteration we need to compute only the first singular vector and there are fast methods to do so \cite{comon1990tracking}. The pair of \eqref{eq:problem_sep1} and \eqref{eq:problem_match1} optimizes and matches a column. Similarly, performing \eqref{eq:problem_sep2} and \eqref{eq:problem_match2} provides us with an optimized row. However, we do not update both of them per each iteration. In fact, we need to perform a column update or a row update in each iteration. It should be determined that which update (column or row) is more efficient in each iteration of the algorithm.  To this aim, first, we choose a random previously selected column and a random previously selected row. Then, we find the best possible replacement column and the best possible replacement row.  Accordingly, we choose whichever that minimizes the cost function more. The best modified column-wise subset is denoted by $\tilde{\mathbb{S}}^c$ and $\tilde{\mathbb{S}}^r$ denotes the best row-wise modified subset. Alg. \ref{alg:jcrss} indicates the steps of TWSP algorithm.  Here, $\dagger$ refers to the Moore–Penrose pseudo-inverse operator. Iterations can be terminated either once CUR decomposition error is saturated or when a maximum number of iterations is reached.

\begin{figure*}[t]
\label{fig:synthesis}
\begin{subfigure}{0.235\textwidth}
\centering
\includegraphics[width=1.45 in]{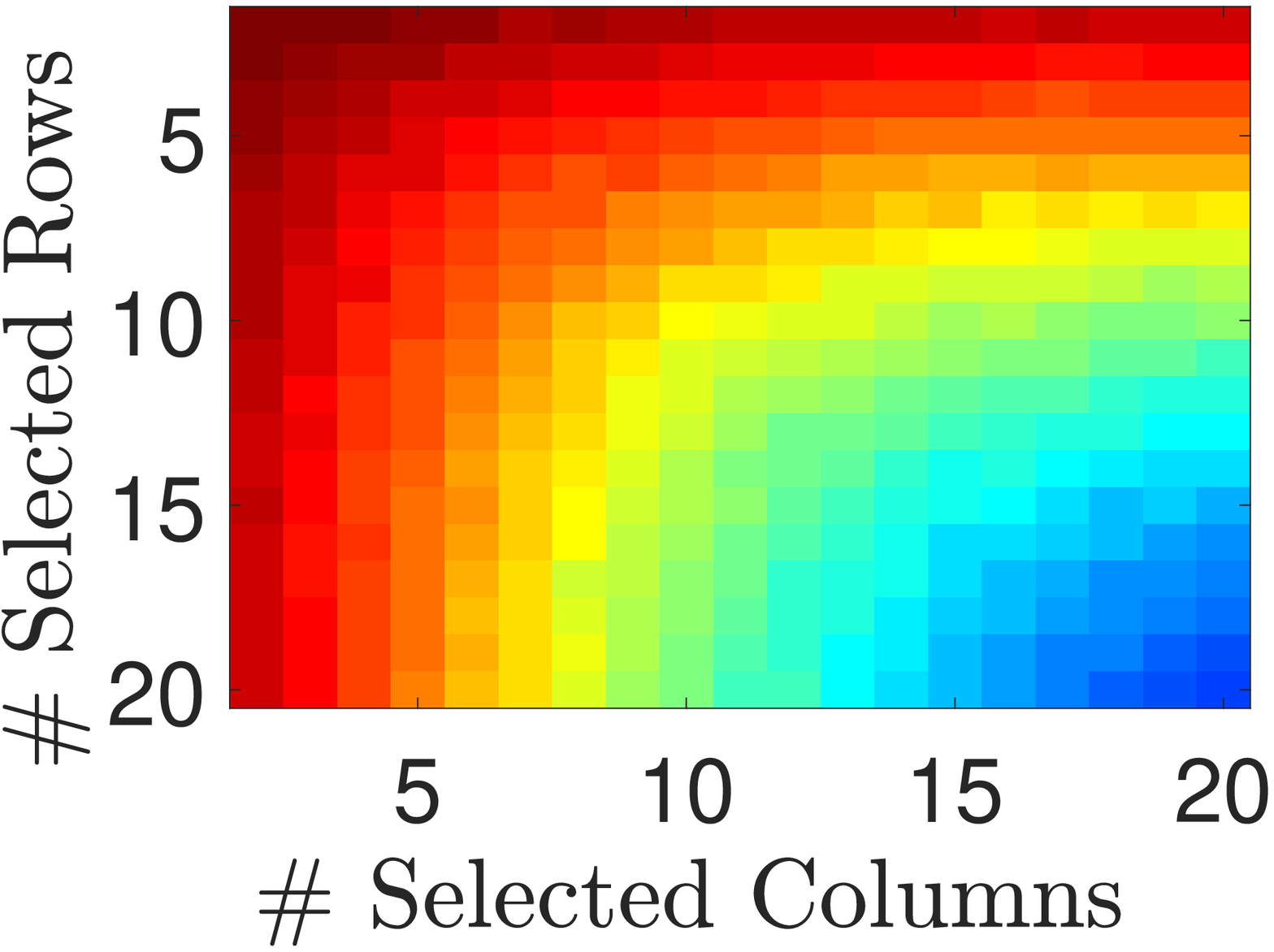}
\footnotesize{\caption{Adaptive CUR \cite{mahoney2009cur}}}
\end{subfigure}
\begin{subfigure}{0.235\textwidth}
\centering
\includegraphics[width=1.45 in]{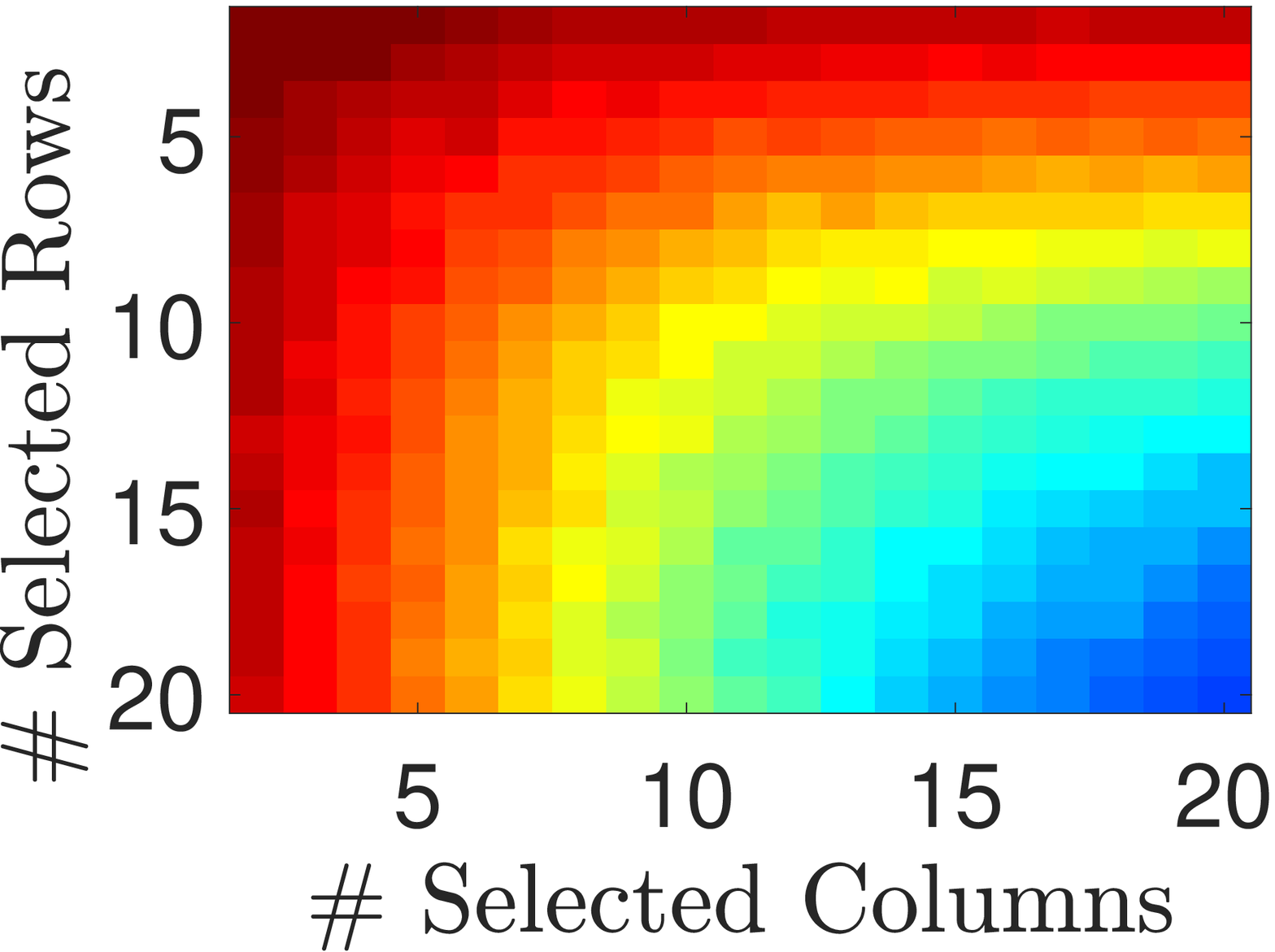}
\footnotesize{\caption{Near-optimal \cite{wang2013improving}}}
\end{subfigure}
\begin{subfigure}{0.235\textwidth}
\centering
\includegraphics[width=1.45 in]{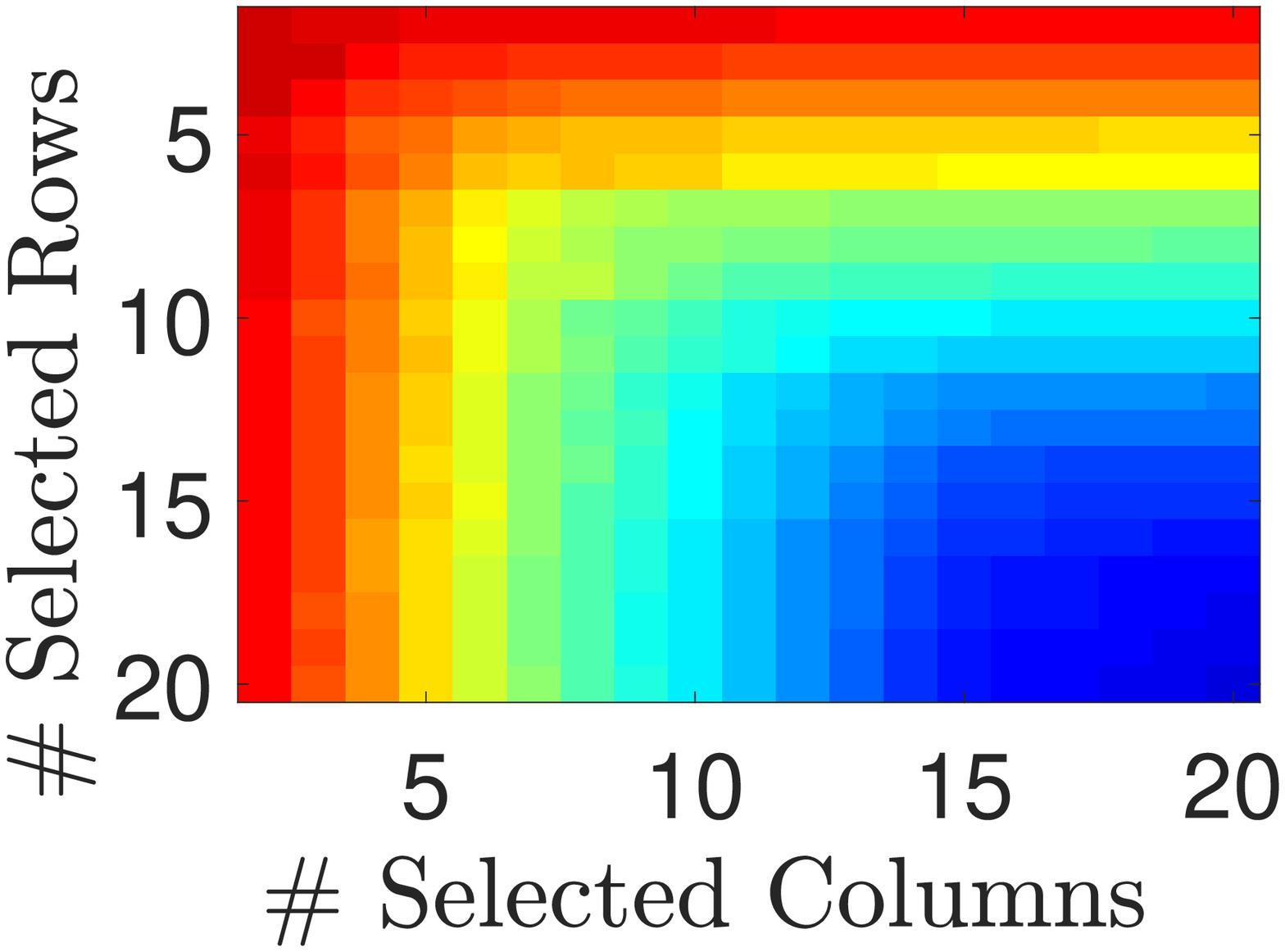}
\footnotesize{\caption{SP \cite{joneidi2020select}}}
\end{subfigure}
\begin{subfigure}{0.255\textwidth}
\centering
\includegraphics[width=1.63 in,height=1.03in]{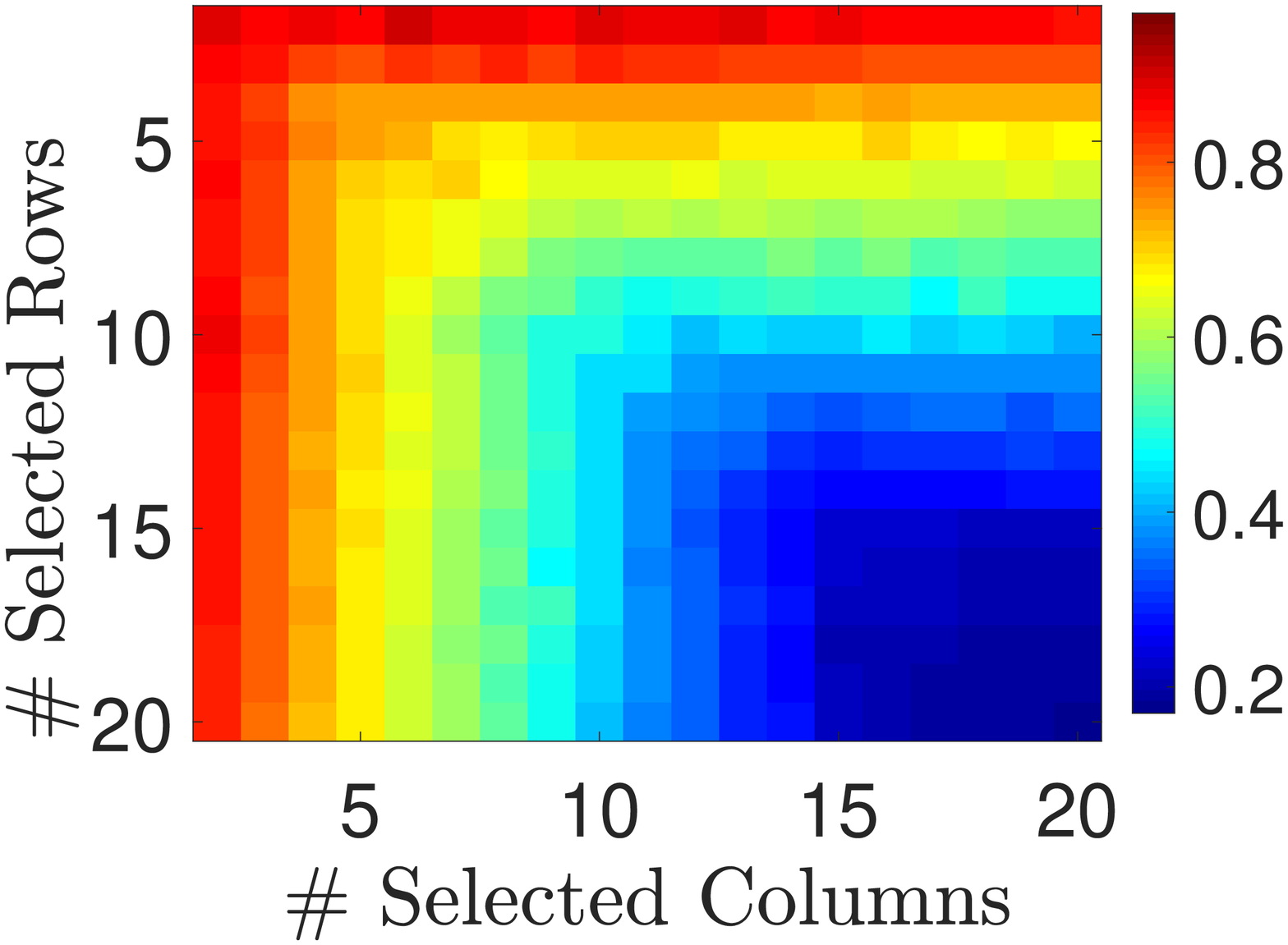}
\footnotesize{\caption{TWSP}}
\end{subfigure}

\footnotesize{\caption{(a)-(d) Performance comparison in terms of the normalized error of CUR decomposition.}}
\vspace{-1mm}
\end{figure*}

\begin{figure}[t]
\vspace{-3mm}
\centering
\includegraphics[width=2.9in, height=1.55in]{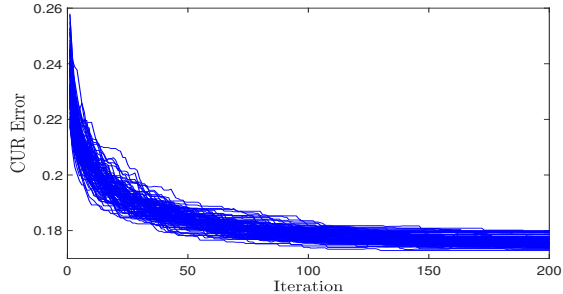}
  \caption{\footnotesize{The convergence behavior of the proposed algorithm w.r.t. the initial condition of selected subset. The initial cost function is corresponding to the initial set which is drawn randomly. The blue curves indicate the optimization path alongside iterations of the TWSP algorithm. Here, $100$ different realizations are studied.}}
  \vspace{-4mm}
\label{fig:init_convrgn}
\end{figure}

The proposed TWSP provides the selected columns and selected rows in order to form matrix $\boldsymbol{C}$ and $\boldsymbol{R}$ in CUR decomposition. It is straightforward to estimate the core matrix $\boldsymbol{U}$. Mathematically,
\vspace{-1mm}
\begin{equation}
\label{eq:core}
    \boldsymbol{U}=\boldsymbol{C}^{\dagger} \boldsymbol{X}\boldsymbol{R}^{\dagger.}
\vspace{-1mm}
\end{equation}
This matrix is a two-way compressed replica of the whole dataset and it contains valuable information in practice as will be discussed in Sec. \ref{subsec:joint}. In general, the number of selected columns may differ from the desired number of rows. Here, $K_1$ refers to the number of columns and $K_2$ points to the number of rows. It is worthwhile to mention that the  complexity order of TWSP is bottle-necked by computational burden for two pseudo-inverses and two singular vectors computation. Thus, the complexity can be expressed as  $\small{O(NK_1^2\!+\!MK_2^2\!+\!MN)}$ per iteration and the algorithm needs $\small{O(\text{max}(K_1,K_2))}$ iterations. Moreover, operator rnd($K$) refers to an integer random number less than or equal to $K$. In the next section, we evaluate the performance of our proposed algorithm.

The CSSP and CUR decomposition problems are NP-hard, i.e., a combinatorial search is required to find the best columns and rows. The proposed TWSP algorithm minimizes the main cost function \eqref{eq:main} in practice. However, there is no theoretical guarantee for convergence of the proposed TWSP algorithm Alg. \ref{alg:jcrss}. 
In order to improve the convergence behavior of TWSP, we evade updating both columns and rows in each iteration. Rather, we prioritize updating  a row or a column depending on which exhibits a smaller projection error, i.e., which is a better minimizer for the cost function per each iteration. The implementation steps of TWSP are summarized in Alg. \ref{alg:jcrss}. As seen in Alg. \ref{alg:jcrss}, the method adds up a revision feature to remedy the greediness associated with consecutive selection by removing samples and optimizing to select a sample for replacement and reiterating this procedure.

\begin{algorithm}[h]
\caption{Two way spectrum pursuit (TWSP)}
\label{alg:jcrss}
\begin{algorithmic}
\small{
\REQUIRE {\small{${\boldsymbol{X}}\!\in\! \mathbb{R}^{N\times M}\!$, $K_1$ and $K_2$.}}\\ \vspace{1.2mm}
\hspace{-3.3mm}\textbf{Output:} $\mathbb{S}^c$ and $\mathbb{S}^r$}. 
\STATE \textbf{Initialization:} \\
 $\mathbb{S}^c\leftarrow$A random subset of $\{1,\ldots,M\}$ with $|\mathbb{S}^c|=K_1$ \\
  $\mathbb{S}^r\leftarrow$A random subset of $\{1,\ldots,N\}$ with $|\mathbb{S}^r|=K_2$ \\ 
 $\{\mathbb{S}^c_k\}_{k=1}^K\!\!\leftarrow\!$ Partition $\mathbb{S}^c$ into $K_1$ subsets.\\
  $\{\mathbb{S}^r_k\}_{k=1}^K\!\!\leftarrow\!$ Partition $\mathbb{S}^r$ into $K_2$ subsets\\
 $i=\text{rnd}(K_1)$ and $j=\text{rnd}(K_2)$ and $\boldsymbol{X}=\boldsymbol{CUR}$ (CUR of $\boldsymbol{X}$)\\ 
 while a stopping criterion is not met \\
 \vspace{+2mm}
  $\qquad \mathbb{S}^c_{\overline{i}}=\mathbb{S}^c\backslash \mathbb{S}^c_i$\\
  \vspace{+1mm}
  $\qquad \boldsymbol{C}_i\leftarrow\;\;\text{remove column}\;i\; \text{in matrix}\;\boldsymbol{C}$\\
  \vspace{+1mm}
  $\qquad \boldsymbol{W}= \boldsymbol{C}_i^{\dagger}\boldsymbol{X}\boldsymbol{R}^{\dagger}$\\
  $\qquad \boldsymbol{E}^c=\boldsymbol{X}-\boldsymbol{C}_i\boldsymbol{WR}\;\;$ {\footnotesize (Null space projection)}\\
  \STATE $\qquad$ $\boldsymbol{c}=\;$find the first left singular vector of $\boldsymbol{E}^c$ \eqref{eq:problem_sep1}\\
\STATE $\qquad$ $\mathbb{S}_i^c\xleftarrow[]{}$  the most correlated column of $\boldsymbol{E}$ with $\boldsymbol{c}$  \eqref{eq:problem_match1}
  \vspace{1mm}
\STATE $\qquad$ $\tilde{\mathbb{S}}^c\xleftarrow[]{} \bigcup_{i'=1}^{K_1} \mathbb{S}^c_{i'}$
  \vspace{1mm}
\STATE $\qquad \boldsymbol{C}=\boldsymbol{X}(:,\tilde{\mathbb{S}}^c)$
  \vspace{1mm}
\STATE $\qquad e^c=\underset{U}{min}\|\boldsymbol{X}-\boldsymbol{C}\boldsymbol{U}\boldsymbol{R}\|_F$\\
  \vspace{3mm}
  $\qquad \mathbb{S}^r_{\overline{j}}=\mathbb{S}^r\backslash \mathbb{S}^r_j$\\
  \vspace{+1mm}
        $\qquad \boldsymbol{R}_j\leftarrow\;\;\text{remove row}\;j\; \text{in matrix}\;\boldsymbol{R}$\\
    $\qquad \boldsymbol{Y}= \boldsymbol{C}^{\dagger}\boldsymbol{X}\boldsymbol{R}_j^{\dagger}$\\
    \vspace{+1mm}
  $\qquad \boldsymbol{E}^r=\boldsymbol{X}-\boldsymbol{CYR}_j \;\;${\footnotesize (Null space projection)}\\
  \STATE $\qquad$ $\boldsymbol{r}=\;$find the first right singular vector of $\boldsymbol{E}^r$ \eqref{eq:problem_sep2} \\
\STATE $\qquad$ $\mathbb{S}_j^r\xleftarrow[]{}$ the most correlated row of $\boldsymbol{E}$ with $\boldsymbol{r}$ \eqref{eq:problem_match2}
\STATE $\qquad$ $\tilde{\mathbb{S}}^r\xleftarrow[]{} \bigcup_{j'=1}^{K_2} \mathbb{S}^r_{j'}$ \vspace{+1mm}
\STATE $\qquad \boldsymbol{R}=\boldsymbol{X}(\tilde{\mathbb{S}}^r,:)$
  \vspace{1mm}
\STATE $\qquad e^r=\underset{U}{min}\|\boldsymbol{X}-\boldsymbol{C}\boldsymbol{U}\boldsymbol{R}\|_F$\\
\vspace{+2mm}
\STATE $\qquad \text{IF }e^c<e^r$
\STATE $\qquad \qquad \mathbb{S}^c\xleftarrow[]{}\tilde{\mathbb{S}}^c$\\
\STATE $\qquad \qquad i=\text{rnd}(K_1)$\\
\STATE $\qquad \text{else}$
\STATE $\qquad \qquad \mathbb{S}^r\xleftarrow[]{}\tilde{\mathbb{S}}^r$\\
\STATE $\qquad \qquad j=\text{rnd}(K_2)$\\
\end{algorithmic}
\end{algorithm}
\vspace{-2mm}
\section{Experimental Results}
\label{sec:exp}
In order to evaluate TWSP on machine-learning tasks in terms of CUR decomposition accuracy, we apply the proposed TWSP on synthetic data as well as three real applications. 
\vspace{-2mm}
\subsection{CUR Decomposition on Synthetic Data}
In order to evaluate the general performance of TWSP, we compared it with the state-of-the-art methods for selecting columns and rows. In this regards, we created a $1000\times2000$ synthetic dataset. The dataset is generated by a rank-$30$ matrix contaminated with random noise. In Fig.~2, we have illustrated the CUR decomposition error for selecting a subset of rows and columns in the range of $2$ to $20$.  The reconstruction error of CUR is normalized by $\|\boldsymbol{X}\|_F^2$. We employ SP  as the state-of-the-art algorithm for column subset selection \cite{joneidi2020select}. We perform SP on the data matrix $\boldsymbol{X}$ and its transpose in order to, respectively, select a subset of columns and rows independently. Then, employing \eqref{eq:core} results in a CUR decomposition. We refer to the algorithm in \cite{mahoney2009cur} as adaptive CUR. 
A more accurate algorithm for solving CUR decomposition results in a bigger blue region in Fig. 2. TWSP exhibits the best performance in this experiment. The convergence behavior of TWSP for this experiment is shown in Fig. 3  for selecting 20 columns and 20 rows. The final solution of the TWSP algorithm depends on the initial selected columns and selected rows. However, regardless of the initial condition, the TWSP algorithm minimizes the cost function of CUR decomposition. TWSP prioritizes in selection of columns or rows such that the largest decrease in the projection error is obtained. It is the main feature to obtain convergence in practice.

\vspace{-2mm}
\subsection{Joint Sensor Selection and Channel Assignment}

\label{subsec:joint}
The output products of CUR decomposition are not limited to a subset of columns and  rows. In some applications, interestingly, matrix $\boldsymbol{U}$ is the most important output of a CUR decomposition. Entry ($i,j$) in $\boldsymbol{U}$ indicates how important the contribution of the $i^{\text{th}}$ column and the $j^{\text{th}}$ row is to reconstruct the whole matrix $\boldsymbol{X}$. This interesting property is utilized for the problem of joint sensor selection and channel assignment in a cognitive radio network. To this aim the exact setup in \cite{zhang2020spectrum} is considered with $900$ grid points and $32$ frequency channels. The received power magnitudes are organized in a $900\times 32$ matrix. 

\begin{figure}[t]
\vspace{-2mm}
\begin{subfigure}{0.48\textwidth}
\centering
\includegraphics[width=3.3 in,height=0.9in]{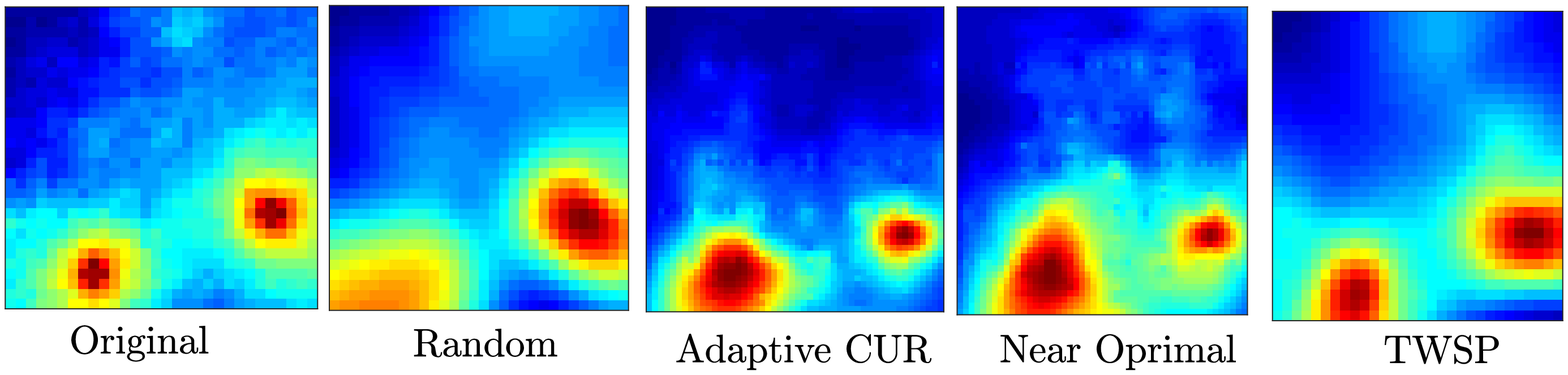}
\end{subfigure}
\begin{subfigure}{0.48\textwidth}
\centering
\includegraphics[width=2.5 in]{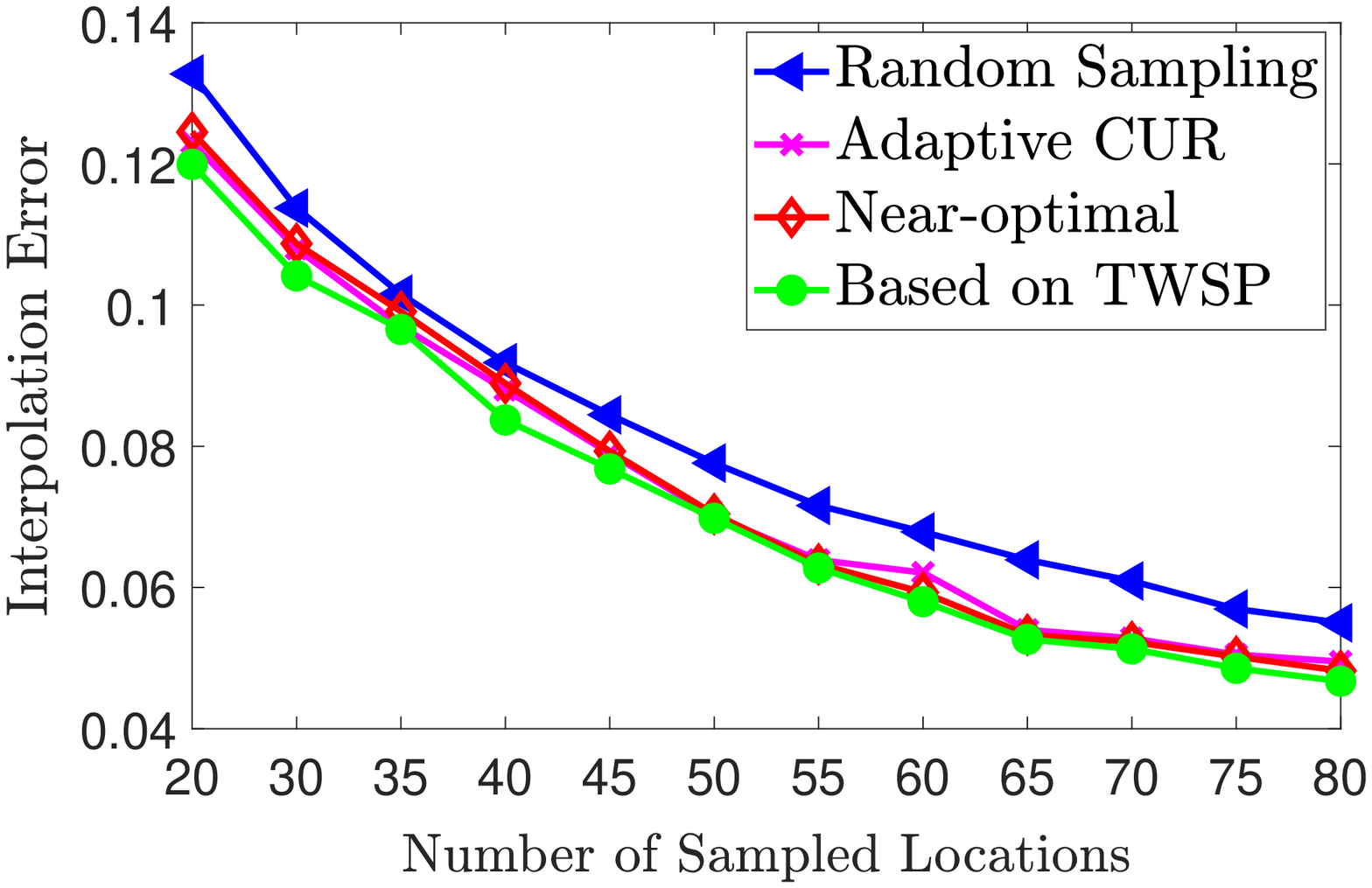}
\end{subfigure}
\vspace{-1mm}
\label{fig:joint_channel}
\vspace{-3mm}
\caption{\footnotesize{The original spectrum map and its comparison with the interpolated map using random sampling and our proposed method. The interpolation error is depicted versus number of sensed locations.}\vspace{-3mm}}
\label{fig:joint_sense}
\end{figure}

The only difference here, is that the uniform sampling pattern of sensing  is replaced by the selection based on the CUR decomposition. Our proposed TWSP algorithm provides a fast and accurate solution for CUR decomposition. We select between $20$ and $80$ locations for spectrum sensing and all  $32$ channels. Each row of matrix $\boldsymbol{U}$  corresponds to a selected location and it has $32$ entries. We are to assign $F$ channels for each selected sensor. In other words, each location does not sense the whole spectrum and only $F$ frequency channels are assigned to each sensor.  The top-$F$  entries in each row with the highest absolute value show the most important channels for the corresponding locations to be sensed.  \\
Fig. \ref{fig:joint_sense} shows the cartography error of spectrum sensing for the conventional random selection as introduced in \cite{zhang2020spectrum} and our proposed optimized joint sensors and channels. For each sampled locations $F=8$ channels out of $32$ channels are sensed. The sampled spectrum map is interpolated using thin plane splines method \cite{ureten2012comparison} for both sampling methods. In addition to visual superiority in reconstruction of the spectrum map, channel assignment based on TWSP provides a better quantitative error as observed in Fig. \ref{fig:joint_sense} 

\subsection{Supervised Sampling}
The proposed TWSP algorithm is an unsupervised data selection algorithm. In supervised settings, labels can infuse information to perform a more viable joint selection \cite{esmaeili2018transduction}. A naive approach is to select representatives from each class independently. However, considering classes jointly is more effective for data reduction. Assume we are given two data classes as $\boldsymbol{X}_1\in \mathbb{R}^{N\times M_1}$ and $\boldsymbol{X}_2\in \mathbb{R}^{N\times M_2}$. The goal is to select $K_1$ samples from class 1 and $K_2$ samples from class 2. To this aim, we propose to construct the cross correlation of two classes as a kernel representation for both classes jointly.  Matrix $\boldsymbol{X}={\boldsymbol{X}_2^T\boldsymbol{X}_1}$ which has $M_1$ columns and $M_2$ rows is fed to TWSP algorithm in order to select $K_1$ columns and $K_2$ rows jointly.

\begin{figure}[!b]
\vspace{-4mm}
\centering
\includegraphics[width=2.8in, height=1.55in]{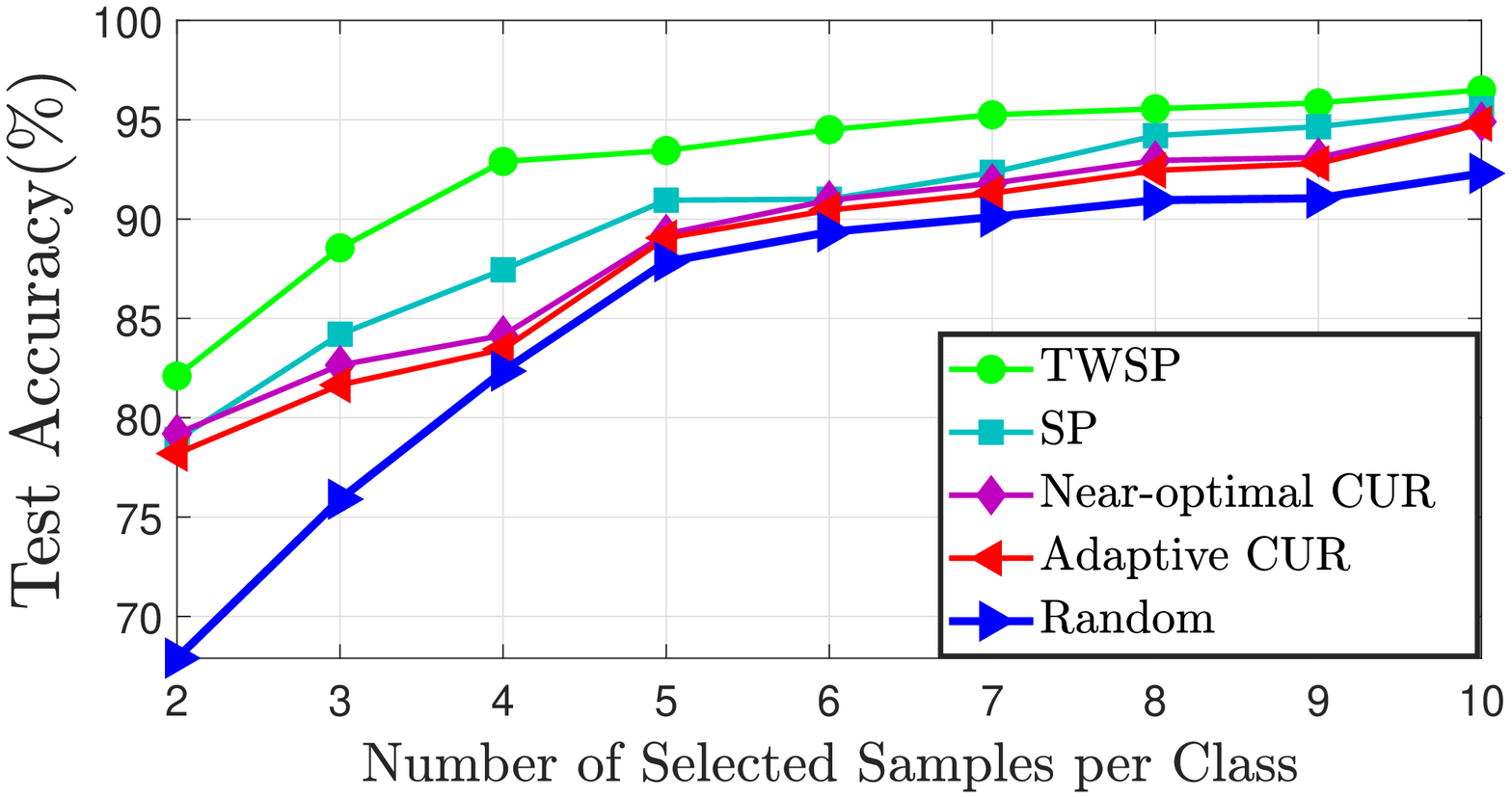}
  \caption{\small{The classification accuracy of a fine-tuned Resnet-18 network using a few selected data per each class.}}
\label{fig:supervised}
\end{figure}

As an initial experiment, supervised sampling is performed on Kaggle cats and dogs dataset. The features are obtained by a trained Resnet-18 deep learning model as explained in \cite{he2016deep}. Three mutually exclusive data subsets for training, validation, and testing are partitioned randomly from $2000$ images of each class. The classification accuracy of $97.5$\% is achieved from a fine-tuned Resnet-18 using the whole training set containing $1000$ samples for each class. Afterwards, samples are selected by applying TWSP on the kernel feature matrix and the Resnet-18 is fine-tuned by using the sampled data. The model's accuracy is compared on the testing set with other sampling methods. Fig. \ref{fig:supervised} shows the performance of selection algorithms for different numbers of representatives per class. Using only two samples from each class, a classification accuracy of $82.3\%$ can be achieved which is more than $15\%$ improvement compared to random selection and more than $5\%$ improvement compared to other competitors.  

\begin{figure}[!t]
\centering
\includegraphics[width=2.9in, height=1.55in]{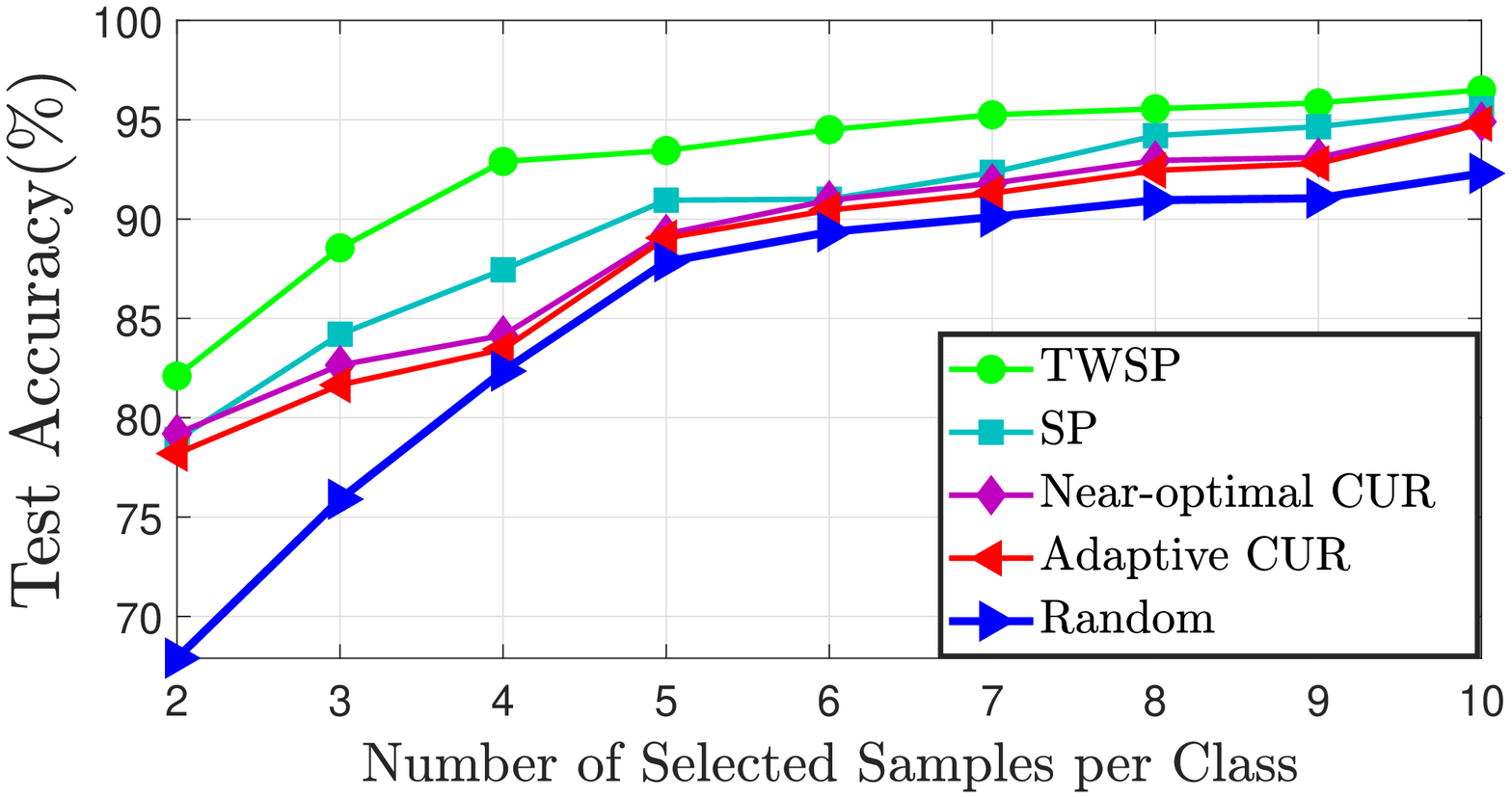}
  \caption{\small{Test accuracy in terms of improvement compared with the test accuracy obtained by random selection.}}
\label{fig:multi_supervised}
\vspace{-3mm}
\end{figure}

We have conducted further experiments to study the effectiveness of the proposed algorithm in the multi-class image classification problem. For this study, we use the Resnet-34 deep learning model pre-trained on CIFAR10 and trained on a subset of ImageNet Dataset comprising of $10$ classes. The classes used for this experiment are Tench, Goldfish, Great white shark, Tiger shark, Hammerhead, Electric ray, Stingray, Cock, Hen and Ostrich. The original training data consists of $1300$ images of each class, and the idea is to use TWSP to select the data samples such that $K$ samples of each class are used for training. 
After training Resnet-34 for 10 classes of CIFAR10, the feature vectors of all the training images are extracted such that a 1300X512 matrix is obtained for each class. Since, TWPS is applicable for the two-class problem, we employ the one-versus-all approach. In other words, a separate cross correlation matrix is generated for each class such that $\boldsymbol{X}_1$ has the dimensions 512X1300 and $\boldsymbol{X}_2$ has the dimensions 512X11700 where $\boldsymbol{X}_1$ represents the feature vector of the class for which samples will be selected while $\boldsymbol{X}_2$ represents the feature vectors of the remaining $9$ classes. Hence, the number of cross correlation matrices is equal to the number of the classes. Then, $K_1$ rows and $K_2$ columns are selected separately by applying TWSP on each cross correlation matrix. This step of generating correlation matrices is different from binary classification where only one correlation matrix was formed. Pre-trained Resnet-34 on CIFAR10 has been refined again by using the sampled data from Imagenet dataset. The model’s accuracy is subsequently compared on the test set with other sampling methods in Fig. \ref{fig:multi_supervised}. As it can be seen, Our proposed TWSP algorithm shows a superior performance comparison to the state of the art methods in sampling.   
\vspace{-1mm}
\subsection{Informative Users/Contents Detection}
Another problem gaining a lot of interest by streaming  service providers is choosing a set of users to decisively reflect their feedbacks about different products. Therefore, it is crucial for such companies to find a subset of users and media products, reviews of which contains information about other users' unknown behavior. 
Each user has a limited scope of interest. 
For example, a user who only loves romance and action genres corresponds to a specific personality that is able to represent a cluster of users accurately. Moreover, such reviews for their areas of interest are more valuable, not reviews for all genres.
Moreover, there exist movies well representing their genres. 

\begin{figure}[!t]
\centering
\includegraphics[width=2.9in, height=1.55in]{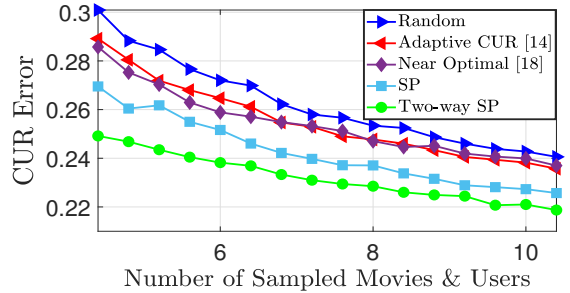}
  \caption{\small{Comparison of the normalized prediction error  with state-of-the-art algorithms obtained by CUR decomposition for simultaneous movies and users subset selection from Netflix dataset.}}
  \vspace{-4mm}
\label{fig:Netflix}
\end{figure}

As a result, there exists a demand for a reliable algorithm to simultaneously choose the most informative subset of users and movies. Such a subset is desirable for streaming companies to the extent that they are willing to give the users incentives to leave comprehensive reviews for certain products. In this regards, we have evaluated our algorithm on Netflix Prize dataset containing $17,770$ movies and $480,189$ users. We have reduced the dataset to $990$ movies and $4,727$ users by considering only movies and users with most reviews. Then, we completed the dataset by Lin et al. method to have a ground truth \cite{lin2010augmented}.
We select a subset of rows and columns (users and media) from the completed dataset. Fig. \ref{fig:Netflix} reveals that TWSP shows the best performance in terms of predicting scores for all users/movies based on a few selected users/movies.

\vspace{-1mm}
\section{Conclusion}
Two-way spectrum pursuit is proposed as an accurate and efficient algorithm for solving CUR decomposition. TWSP can be employed for joint selection of columns and rows such that their outer products is able to reconstruct the whole matrix as accurate as possible. Some applications of the proposed algorithm are presented to show the efficacy of the proposed method. However, they are not limited to the mentioned applications. Moreover, the proposed algorithm can be extended to $n$-way spectrum pursuit for efficient tensor subset selection. 

\bibliographystyle{plain}
\bibliography{main}
\end{document}